\pdfoutput=1
\documentclass[11pt]{article}

\usepackage{naacl2021}

\usepackage[utf8]{inputenc}
\usepackage{latexsym}
\usepackage[T1]{fontenc}
\usepackage{microtype}

\usepackage{latexsym}
\usepackage{caption}
\usepackage{url}
\usepackage[titletoc,title]{appendix}
\usepackage{soul}
\usepackage{tabularx}
\usepackage{caption}
\usepackage{amsthm}
\theoremstyle{definition}

\theoremstyle{remark}

\usepackage{amssymb}
\usepackage{url}
\usepackage{makecell}
\usepackage{graphicx}
\usepackage{kotex}
\usepackage{array,multirow}
\usepackage{booktabs}
\usepackage{CJKutf8}
\usepackage{makecell}
\RequirePackage{filecontents}
\usepackage[strict]{changepage}
\usepackage{amsmath}
\usepackage{subcaption}
\newcolumntype{b}{X}
\newcolumntype{t}{>{\hsize=.15\hsize}X}
\newcolumntype{y}{>{\hsize=.25\hsize}X}
\newcolumntype{s}{>{\hsize=.85\hsize}X}
\newcolumntype{a}{>{\hsize=1.05\hsize}X}
\newcolumntype{u}{>{\hsize=1.2\hsize}X}
\newcolumntype{l}{>{\hsize=1.5\hsize}X}
\newcolumntype{w}{>{\hsize=2\hsize}X}
\newcolumntype{v}{>{\hsize=2.5\hsize}X}
\newcolumntype{z}{>{\centering\hsize=6\hsize}X}
\usepackage{etoolbox}
\usepackage{textcomp}
\usepackage{times}

\title{Family of Origin and Family of Choice: \\
  Massively Parallel Lexiconized Iterative Pretraining \\
  for Severely Low Resource Text-based Translation}

\author{
  Zhong Zhou\\
  {\small Carnegie Mellon University}\\
  {\small \tt zhongzhou@cmu.edu}
  \\\And
  Alex Waibel\\
  {\small Carnegie Mellon University}\\
  {\small \tt alex@waibel.com }
  }

\begin{document}
\maketitle
\begin{abstract}
We translate a closed text that is known in advance into
a severely low resource language by leveraging massive
source parallelism. In other words,
given a text in 124 source languages,
we translate it into a severely low resource language
using only $\sim$1,000 lines of low resource data without any external help.
Firstly, we propose a systematic method to rank
and choose source languages that are close to
the low resource language. 
We call the linguistic definition of
language family \textit{Family of Origin} (FAMO), and
we call the empirical definition of higher-ranked
languages using our metrics
\textit{Family of Choice} (FAMC).
Secondly, we build an
\textit{Iteratively Pretrained Multilingual Order-preserving Lexiconized Transformer} (IPML)
to train on $\sim$1,000 lines ($\sim$3.5\%) of low resource data.
To translate named entities correctly,
we build a massive lexicon table for
2,939 Bible named entities in 124 source languages,
and include many that occur once and
covers more than 66 severely low resource languages.
Moreover, we also build a novel method of
combining translations from different
source languages into one. Using English
as a hypothetical low resource language,
we get a +23.9 BLEU increase over a multilingual
baseline, and a +10.3 BLEU increase over our asymmetric baseline
in the Bible dataset. We get a 42.8 BLEU score for Portuguese-English translation
on the medical EMEA dataset. 
We also have good results for a real severely low resource
Mayan language, Eastern Pokomchi.  
\end{abstract}


\section{Introduction} \label{introduction}
We translate a closed text that is known in advance into
a severely low resource language by leveraging massive
source parallelism. In other words, we aim to translate well under
three constraints: having severely small training data
in the new target low resource language,
having massive source language parallelism,
having the same closed text across
all languages. Generalization to other texts
is preferable but not necessary
in the goal of producing high quality
translation of the closed text.

2020 is the year that we started the
life-saving hand washing practice globally.
Applications like translating water, sanitation,
and hygiene (WASH) guidelines into severely
low resource languages are very impactful in
tribes like those in Papua New Guinea with
839 living languages \cite{gordon2005ethnologue, simons2017ethnologue}.
Translating humanitarian texts like WASH guidelines
with scarce data and expert help is key \cite{bird2020decolonising}. 
\begin{table}[t]
  \small
  \centering
  \begin{tabularx}{\columnwidth}{ss|ss} 
    \toprule
    \multicolumn{2}{w|}{Eastern Pokomchi} & \multicolumn{2}{w}{English} \\ 
    \midrule
    \textit{FAMD} & \textit{FAMP} & \textit{FAMD} & \textit{FAMP} \\ 
    \midrule
    Chuj* & Dadibi & Danish* & Dutch* \\ 
    Cakchiquel* & Thai & Norwegian* & Afrikaans* \\ 
    Guajajara* & Gumatj & Italian & Norwegian* \\ 
    Toba & Navajo & Afrikaans* & German* \\
    Myanmar & Cakchiquel* & Dutch* & Danish* \\
    Slovenský & Kanjobal & Portuguese & Spanish \\ 
    Latin & Guajajara* & French & Frisian* \\
    Ilokano & Mam* & German* & Italian \\ 
    Norwegian & Kim & Marshallese & French \\ 
    Russian & Chuj* & Frisian* & Portuguese \\ 
    \bottomrule
  \end{tabularx}
    \caption{Top ten languages closest to Eastern Pokomchi (left)
      and English (right) in ranking 124 source languages. \textit{FAMD} and \textit{FAMP} are two
      constructions of Family of Choice (\textit{FAMC}) by distortion and performance metrics respectively. 
      All are trained on $\sim$1,000 lines. 
      We star those in Family of Origin.}
    \label{table:ranking}
\end{table}

We focus on five challenges that are not addressed previously.
Most multilingual transformer works that translate
into low resource language
limit their training
data to available data in the same
or close-by language families 
or the researchers' intuitive discretion;
and are mostly limited to less than 30 languages
\cite{gu2018universal, zhong2018massively, zhu2020language}.
Instead, we examine ways to pick useful
source languages from 124 source languages in a principled fashion. 
Secondly, most works require at least
4,000 lines of low resource data \cite{lin2020pre, qi2018and, zhong2018massively};
we use 
only $\sim$1,000 lines of low resource data to simulate real-life situation
of having extremely small seed target translation.
Thirdly, many works use rich resource languages as hypothetical
low resource languages. 
Moreover, most works do not treat named entities separately; 
we add an order-preserving lexiconized component
for more accurate translation of named
entities. Finally, many multilingual works present final
results as sets of translations from all source languages;
we build a novel method to combine all translations into one. 

We have five contributions. Firstly, we rank the
124 source languages to determine their
closeness to the low resource language and choose the top few. 
We call the linguistic definition of
language family \textit{Family of Origin} (FAMO), and
we call the empirical definition of higher-ranked 
languages using our metrics 
\textit{Family of Choice} (FAMC).
They often overlap, but may not coincide. 

Secondly, we build an
\textit{Iteratively Pretrained Multilingual Order-preserving Lexiconized Transformer} (IPML)
training on $\sim$1,000 lines
of low resource data. 
Using iterative pretraining, 
we get a +23.9 BLEU increase over
a multilingual order-preserving lexiconized transformer
baseline (MLc)
using English as a hypothetical low resource
language, and a +10.3 BLEU increase over our asymmetric baseline. 
Training with the low resource
language on both the source and target sides
boosts translation into the target side.
Training on 1,093 lines from the book of Luke, 
we reach a 23.7 BLEU score testing on 30,022 lines of Bible.
We have a 42.8 BLEU score for Portuguese-English translation
on the medical EMEA dataset.

Thirdly, we use a real-life severely
low resource Mayan language, Eastern Pokomchi,
a Class 0 language \cite{joshi2020state}
as one of our experiment
setups. In addition, we also use English
as a hypothetical low resource language for easy 
evaluation. 

We also add an order-preserving lexiconized
component to translate named entities well. To
solve the variable-binding
problem to distinguish ``Ian calls Yi''
from ``Yi calls Ian'' \cite{fodor1988connectionism, graves2014neural, zhong2018massively},
we build a lexicon table for
2,939 Bible named entities in 124 source languages
including more than 66 severely low resource languages. 

Finally, we combine translations from all source
languages by using a novel method. For every sentence,
we find the translation that is closest to
the translation cluster center.
The expectated BLEU score of our combined translation
is higher than translation from any of the individual source. 

\section{Related Works} \label{relatedwork}
\subsection{Information Dissemination}
Interactive Natural Language Processing (NLP)
systems are classified into information assimilation,
dissemination, and dialogue
\cite{bird2020decolonising, ranzato2015sequence, waibel2008spoken}. 
\textit{Information assimilation} involves information
flow from low resource to rich resource
language communities while \textit{information dissemination}
involves information flow from rich resource to low resource language communities.
Taken together, they allow \textit{dialogue} and interaction of
different groups at eye level.
Most work on information assimilation
\cite{berard2020multilingual, earle2012twitter, brownstein2008surveillance}. 
Few work on dissemination due to small data,
less funding, few experts and limited writing system
\cite{ostling2017neural, zoph2016transfer, anastasopoulos2017spoken, adams2017cross, bansal2017weakly}.

\subsection{Machine Polyglotism and Pretraining}
Recent research on machine polyglotism involves
training machines to be adept in many languages
by adding language labels in the training data
with a single attention 
\cite{johnson2017google, ha2016toward, firat2016multi, gillick2016multilingual, zhou2018paraphrases}. Some explores data symmetry \cite{freitag2020complete, birch2008predicting, lin2019choosing}.
Zero-shot translation in severely low resource settings
exploits the massive multilinguality, cross-lingual
transfer, pretraining, iterative back-translation
and freezing subnetworks
\cite{lauscher2020zero, nooralahzadeh2020zero, wang2020negative, li2020learn, pfeiffer2020mad, baziotis2020language, chronopoulou2020reusing, lin2020pre, thompson2018freezing, luong2014addressing, wei2020iterative, dou2020dynamic}.

\subsection{Linguistic Distance}
To construct linguistic distances \cite{hajic2000machine, oncevay2020bridging}, 
some explore  
typological distance
\cite{chowdhury2020understanding, rama2012good, pienemann2005processability, svalberg1998english, hansen2012beginning, comrie2005world}, 
lexical distance 
\cite{huang2007towards},
Levenshtein distance and Jaccard distance
\cite{serva2008indo, holman2008advances, adebara2020translating},
sonority distance \cite{parker2012sonority} 
and spectral distance \cite{dubossarsky2020secret}.

\section{Methodology}\label{method}

\subsection{Multilingual Order-preserving Lexiconized Transformer}
\subsubsection{Multilingual Transformer}
In training, each sentence is labeled with the source and target language label. For example, if we translate from Chuj (``ca'') to Cakchiquel (``ck''),
each source sentence is tagged with 
\texttt{\_\_opt\_src\_ca \_\_opt\_tgt\_ck}. A sample source
sentence is
``\texttt{\_\_opt\_src\_ca \_\_opt\_tgt\_ck} Tec'b'ejec e b'a mach ex tzeyac'och Jehová yipoc e c'ool''.

We train on Geforce RTX 2080 Ti
using $\sim$100 million parameters,
a 6-layer encoder and a 6-layer decoder
that are powered by 512 hidden states,
8 attention heads, 512 word vector size,
a dropout of 0.1, an attention
dropout of 0.1, 2,048 hidden transformer feed-forward units, a
batch size of 6,000, ``adam'' optimizer, ``noam'' decay method,
and a label smoothing of 0.1 and
a learning rate of 2.5 on OpenNMT \cite{klein2017opennmt, vaswani2017attention}.
After 190,000 steps, we validate based on BLEU score
with early stopping patience of 5. 

\subsubsection{Star Versus Complete Configuration}
\begin{figure}
  \centering
  \begin{subfigure}[b]{0.5\textwidth}
  \centering
  \includegraphics[width=0.9\linewidth]{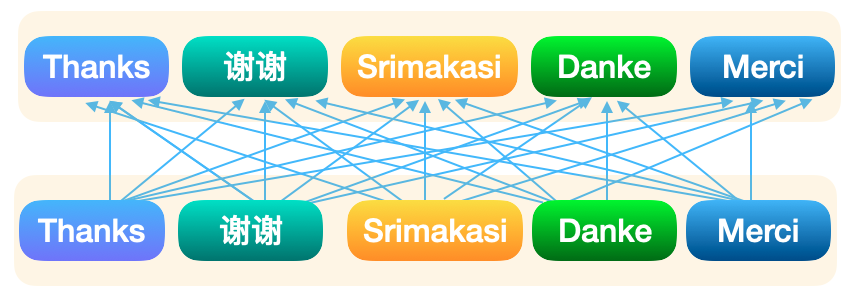}
   \caption{Complete graph configuration}
   \label{fig:complete}
\end{subfigure}
\begin{subfigure}[b]{0.5\textwidth}
  \centering
  \includegraphics[width=0.9\linewidth]{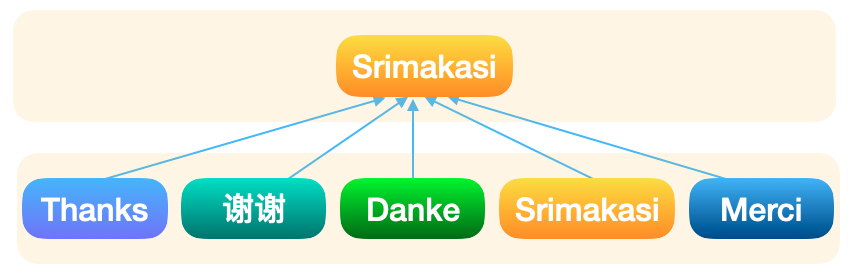}
  \caption{Star graph configuration}
  \label{fig:star}
\end{subfigure}
\caption[Two configurations of translation paths]{(a) Complete
  graph configuration of translation paths (Many-to-many)
  in an example of multilingual translation. (b)
  Star configuration of translation paths (Many-to-one) using Indonesian
  as the low resource example.}
\label{fig:config}
\end{figure}

We show two configurations of translation paths
in Figure~\ref{fig:config}: \textit{star}
graph (multi-source single-target) configuration and
\textit{complete} graph (multi-source multi-target) configuration.
The complete configuration data
increases quadratically with the number of languages 
while the star configuration data increases linearly. 

\subsubsection{Order-preserving Lexiconized transformer}
The variable binding problem issue is difficult in severely
low resource scenario; most neural models cannot distinguish
the subject and the object of a
simple sentence like ``Fatma asks her sister Wati to
call Yi, the brother of Andika'',
especially when all named entities appear
once or never appear in training
\cite{fodor1988connectionism, graves2014neural}.
Recently, researchers use order-preserving lexiconized Neural
Machine Translation models where named entities are sequentially tagged
in a sentence as \texttt{\_\_NE}s  
\cite{zhong2018massively}. The previous example becomes 
``\texttt{\_\_NE0} asks her sister \texttt{\_\_NE1} to                         
call \texttt{\_\_NE2}, the brother of \texttt{\_\_NE3}''.

This method works under the assumption
of translating a closed text known in advance. 
Its success relies on good coverage of named entities.
To cover many named entities,
we build on existing research literature \cite{wu2018creating, zhong2018massively}
to construct a massively parallel 
lexicon	table that covers 2,939 named entities across 124
languages in our Bible database. Our lexicon table
is an expansion of the existing literature that covers 1,129 named
entities \cite{wu2018creating}. We add in 
1,810 named entities that are in the extreme end of the tail
occurring only once. We also include 66 more real-life severely low resource
languages. 

For every sentence pair,
we build a target named entity decoding dictionary
by using all target
lexicons from the lexicon table
that match with those in the
source sentence. In severely low resource setting,
our sequence tagging is larged based on
dictionary look-up; we also include lexicons
that are not in the dictionary but have small
edit distances with the source lexicons.
In evaluation, we
replace all the ordered \texttt{\_\_NE}s
using the target decoding dictionary
to obtain our final translation.

Let us translate ``Fatma asks her sister Wati to call Yi, the brother of Andika''
to Chinese and German. 
Our tagged source sentence that translates to Chinese is
``\texttt{\_\_opt\_src\_en} \texttt{\_\_opt\_tgt\_zh} \texttt{\_\_NE0}
asks her sister \texttt{\_\_NE1} to
call \texttt{\_\_NE2}, the brother of \texttt{\_\_NE3}'';
and we use \texttt{\_\_opt\_tgt\_de} for German.
The source dictionary is ``\texttt{\_\_NE0}: Fatma,
\texttt{\_\_NE1}: Wati, \texttt{\_\_NE2}: Yi,
\texttt{\_\_NE3}: Andika'' and we create the target dictionaries.
The Chinese output is ``\texttt{\_\_NE0}\begin{CJK*}{UTF8}{gbsn}叫她的姐妹\end{CJK*}\texttt{\_\_NE1}\begin{CJK*}{UTF8}{gbsn}去打电话给\end{CJK*}\texttt{\_\_NE3}\begin{CJK*}{UTF8}{gbsn}的兄弟\end{CJK*}\texttt{\_\_NE2}''
      and the German output is ``\texttt{\_\_NE0} bittet ihre Schwester \texttt{\_\_NE1} darum, \texttt{\_\_NE2}, den Bruder \texttt{\_\_NE3}, anzurufen''.
      We decode the named entities to get final translations. 

\subsection{Ranking Source Languages} \label{pathway} \label{dataranking} 
Existing works on 
translation from multiple source
languages into a single low resource language
usually have at most 30 source
languages \cite{gu2018universal, zhong2018massively, zhu2020language}.
They are limited within the same or close-by
language families, or those with available data, or those
chosen based on the researchers' intuitive discretion. 
Instead, we examine ways to pick useful
source languages in a principled fashion
motivated by cross-lingual impacts
and similarities
\cite{shoemark2016towards, sapir1921languages, odlin1989language, cenoz2001effect, toral2018level, de1997pursuit, hermans2003cross, specia2016shared}.
We find that using many languages that are distant to the target
low resource language may produce marginal improvements,
if not negative impact.
Indeed, existing literature on zero-shot translation
also suffers from the limitation of linguistic distance between the
source languages and the target language
\cite{lauscher2020zero, lin2020pre, pfeiffer2020mad}.
We therefore rank and select the top few source
languages that are closer to the target low resource
language using the two metrics below.

We rank source languages according to their closeness to the
low resource language. We construct
the Family of Choice (FAMC) by comparing different ways of
ranking linguistic distances
empirically based on the small low resource data.

Let $S_{s}$ and $S_{t}$ be the source and target
sentences, let $L_{s}$ be the source length,
let $P(S_{t} = s_{t} | s_{s}, l_{s})$ be the alignment
probability, let $F_{s}$ be the fertility of how many target
words a source word is aligned to, let $D_{t}$ be the distortion based on the fixed distance-based reordering model \cite{koehn2009statistical}. 

We first construct a word-replacement model based on aligning
the small amount of target
low resource data with that of each source language
using \texttt{fast\_align} \cite{dyer2013simple}.
We replace every source word with the most probable
target word according to the product of the alignment 
probability and the probability of fertility equalling one and distortion equalling zero  
$P(F_{s} = 1, D_{t} = 0 | s_{t}, s_{s}, l_{s})$.
We choose a simple word-replacement model because 
we aim to work with around 1,000 lines of low resource
data. For fast and efficient ranking on such small data, a word-replacement
model suits our purpose. 

We use two alternatives to create our FAMCs.
Our distortion measure is the probability of distortion equalling zero, 
$P(D_{t} = 0 | s_{t}, s_{s}, l_{s})$, 
aggregated over all words in a source language.
We use the distortion measure to rank
the source languages and obtain the distortion-based FAMC (\textit{FAMD});
we
use the translation BLEU scores of the
word-replacement model as another alternative
to build the performance-based FAMC (\textit{FAMP}).  
In Table~\ref{table:ranking}, we list the top ten
languages in FAMD and
FAMP for Eastern Pokomchi
and English. We use both alternatives to build FAMCs. 

To prepare for transformer training, we choose
the top ten languages neighboring our target low resource
language in FAMD and FAMP.
We choose ten because
existing literature shows that
training with ten languages from two neighboring language
families is sufficient in producing
quality translation through cross-lingual transfer
\cite{zhong2018massively}. Since for
some low resource languages, there may not be
ten languages in FAMO in our database, we add
languages from neighboring families to make
an expanded list denoted by \textit{FAMO$^+$}. 

\subsection{Iterative Pretraining}
We have two stages of pretraining using
multilingual 
order-preserving lexiconized transformer
on the complete and the star configuration.
We design iterative pretraining on symmetric data to
address catastrophic
forgetting that is common in training \cite{french1999catastrophic, kirkpatrick2017overcoming}.

\begin{table*}[]
    \small
    \centering
      \begin{tabularx}{\textwidth}{XXX}
      \toprule
      Source Sentence & IPML Translation & Reference \\
      \midrule
      En terwyl Hy langs die see van Galiléa loop, sien Hy Simon en Andréas, sy broer, besig om 'n net in die see uit te gooi; want hulle was vissers.
      &
      And as He drew near to the lake of Galilee, He Simon saw Andrew, and his brother, lying in the lake, for they were fishermen.
      &
      And walking along beside the Sea of Galilee, He saw Simon and his brother Andrew casting a small net in the sea; for they were fishers.
      \\
      \midrule
      En toe Hy daarvandaan 'n bietjie verder gaan, sien Hy Jakobus, die seun van Sebedéüs, en Johannes, sy broer, wat besig was om die nette in die skuit heel te maak.
      &
      And being in a distance, He saw James, the son of Zebedee, and John, his brother. who kept the nets in the boat.
      &
      And going forward from there a little, He saw James the son of Zebedee, and his brother John. And they were in the boat mending the nets.
      \\
      \midrule
      En verder Jakobus, die seun van Sebedéüs, en Johannes, die broer van Jakobus- aan hulle het Hy die bynaam Boanérges gegee, dit is, seuns van die donder-
      &
      And James the son of Zebedee, and John the brother of James; and He gave to them the name, which is called Boanerges, being of the voice.
      &
      And on James the son of Zebedee, and John the brother of James, He put on them the names Boanerges, which is, Sons of Thunder.
      \\
      \bottomrule
    \end{tabularx}
      \caption{Examples of Iteratively Pretrained Multilingual
        Order-preserving Lexiconized Transformer (IPML) translation
        from Afrikaans to English as a hypothetical low resource language using \textit{FAMP}.
        We train on only 1,093 lines of English data. }
    \label{table:englishQualitative}
  \end{table*}

\subsubsection{Stage 1: Pretraining on Neighbors}\label{step1}
Firstly, we pretrain on the complete
graph configuration of translation paths using the
top ten languages neighboring our target low resource
language in FAMD, 
FAMP, and FAMO$^+$ respectively. Low resource data is excluded in training. 

We use the multilingual order-preserving
lexiconized transformer. Our vocabulary is the combination of the
vocabulary for the top ten languages together with the low resource vocabulary built from the $\sim$1,000 lines.
The final model can translate from any
of the ten languages to each other.

\subsubsection{Stage 2: Adding Low Resource Data} \label{step2}
We include the low resource data
in the second stage of training. Since the low resource
data covers $\sim3.5\%$ of the text while all the source
languages cover the whole text, the data is highly asymmetric.
To create symmetric data, we align the low resource
data with the subset of data from all source languages.
As a result, all source languages in the second stage of training
have $\sim3.5\%$ of the text that is aligned with the
low resource data. We therefore
create a complete graph configuration of training
paths using all the eleven languages.

Using the pretrained model from the previous stage,
we train on the complete
graph configuration of translation paths from all eleven
languages including our low resource
language. The vocabulary used is the same as before.
We employ the multilingual order-preserving lexiconized transformer for pretraining.
The final model can translate from any of the eleven languages to each other.

\subsection{Final Training}\label{step3}
Finally, we focus on translating into the low resource
language.
We use the symmetric data built 
from the second stage of pretraining.
However, instead of using the complete 
configuration, we use the
star configuration of translation
paths from the all source languages to the
low resource language. All languages
have $\sim3.5\%$ of the text. 

Using the pretrained model from the second stage,
we employ the multilingual order-preserving
lexiconized transformer on the star
graph configuration. We use the same vocabulary
as before. 
The final trained model can
translate from any
of the ten source languages to the low resource language.
Using the lexicon dictionaries, we decode the named
entities and obtain our final translations. 

\subsection{Combination of Translations}\label{step4}
We have multiple translations, one per each source language.
Combining all translations is
useful for both potential post-editting works and
systematic comparison of different
experiments especially when the sets of the source languages differ.

Our combination method assumes that we have the same text in
all source languages. For each sentence, we form a cluster of
translations from all source languages into the low resource
language. Our goal is to find the translation that is closest
to the center of the cluster. We rank all translations according to how
centered this translation is with respect to other sentences by
summing all its similarities to the rest. The highest score is the
closest to the cluster center. We take the most centered
translation for each sentence and output our combined result.
The expectated BLEU score of our combined translation
is higher than translation from	any of the individual
source language.

\section{Data}\label{data}
We use the Bible dataset and the medical
EMEA dataset \cite{mayer2014creating, tiedemann2012parallel}.
EMEA dataset is from the European Medicines Agency and
contains a lot of medical information that may be beneficial to the
low resource communities.  
Our method can be
applied to other datasets like WASH guidelines.

For the Bible dataset,
we use 124 source languages with 31,103 lines
of data and
a target low resource language with $\sim$1,000 lines ($\sim$3.5\%) of data.
We have two setups for the
target low resource language.
One uses Eastern Pokomchi,
a Mayan language; 
the other uses English
as a hypothetical low resource language.
We train
on only $\sim$1,000 lines of low resource data
from the book of Luke and test on the 678 lines
from the book of Mark. Mark is
topically similar to Luke, but is written by a different author. 
For the first stage of pretraining, we use
80\%, 10\%, 10\% split for training, validation
and testing. For the second stage onwards, we
use 95\%, 5\% split of Luke for training and validation,
and 100\% of Mark for testing. 

Eastern Pokomchi
is Mayan, and English is Germanic. Since our database does not
have ten members of each family,
we use FAMO$^+$, the expanded version of FAMO. 
For English, we include five Germanic
languages and five Romance languages in FAMO$^+$;
for Eastern Pokomchi, we include five Mayan languages
and five Amerindian languages in FAMO$^+$. 
The Amerindian family is broadly
believed to be close to the Mayan family
by the linguistic community. 

We construct
FAMCs by comparing different ways of
ranking linguistic distances
empirically based on $\sim$1,000 lines of training data.
In Table~\ref{table:ranking}, we list the top ten
languages for Eastern Pokomchi
and English in FAMD and
FAMP respectively.

To imitate the real-life situation of having small
seed target translation data, 
we choose to use $\sim$1,000 lines ($\sim$3.5\%) of
low resource data. We also include Eastern Pokomchi
in addition to using English as a hypothetical low resource language.
Though data size
can be constrained to mimic severely low resource
scenarios, much implicit information is still used for
the hypothetical low resource language
that is actually rich resource.
For example, implicit information like
English is Germanic is often used.
For real low resource scenarios, the family information may have yet to be
determined; the neighboring languages may be unknown, and if they are known, they are highly likely to be low resource too. We thus use Eastern Pokomchi as our real-life severely low resouce language.

In addition to the Bible dataset, we work with the
medical EMEA dataset \cite{tiedemann2012parallel}. Using
English as a hypothetical language, we train on randomly sampled 
1,093 lines, and test on 678 lines of data. 
Since there are only 9 languages in Germanic and
Romance families in EMEA dataset, we include
a slavic language Polish in our FAMO$^+$ for experiments.

The EMEA dataset is less than ideal comparing with the Bible dataset.
The Bible dataset contains the same text for all source languages; however, the EMEA dataset does not
contain the same text. It is
built from similar documents but
has different parallel data for
each language pair.
Therefore, during test time, we do not
combine the translations from various source languages
in the EMEA dataset.

 \begin{table}[t]
  \centering
  \small
  \begin{tabular}{ p{1.55cm}|p{0.52cm}p{0.52cm}p{0.52cm}p{0.52cm}p{0.52cm}p{0.52cm} }
    \toprule
    Experiments & IPML & MLc & MLs & PMLc & PMLs & AML\\
    \midrule
    Pretrained & \checkmark & &  & \checkmark & \checkmark &  \\
    Iterative & \checkmark & & & & & \\
    Lexiconized & \checkmark & \checkmark & \checkmark & \checkmark & \checkmark & \checkmark \\
    Symmetrical & \checkmark & \checkmark & \checkmark & \checkmark & \checkmark \\
    Star & \checkmark & & \checkmark & & \checkmark & \\
    Complete & \checkmark & \checkmark & & \checkmark & & \checkmark \\
    \midrule
    Combined & 37.3 & 13.4 & 14.7 & 34.7 & 35.7 & 27.0 \\
    \midrule 
    German & 35.0 & 11.6 & 12.3 & 33.3 & 34.5 & 25.4 \\
    Danish & 36.0 & 12.5 & 12.4 & 33.3 & 34.2 & 26.2 \\
    Dutch & 35.6 & 11.5 & 11.1 & 32.3 & 33.7 & 25.0 \\
    Norwegian & 35.7 & 12.3 & 12.0 & 33.2 & 34.1 & 25.8 \\
    Swedish & 34.5 & 11.8 & 12.4 & 32.3 & 33.4 & 24.9 \\
    Spanish & 36.4 & 11.7 & 11.8 & 34.1 & 35.0 & 26.2 \\
    French & 35.3 & 10.8 & 10.8 & 33.1 & 34.0 & 25.8 \\
    Italian & 35.9 & 11.7 & 11.7 & 34.3 & 34.5 & 26.1 \\
    Portuguese & 31.5 & 9.6 & 10.1 & 30.0 & 30.4 & 23.1 \\
    Romanian & 34.6 & 11.3 & 12.1 & 32.3 & 33.2 & 25.0 \\
    \bottomrule
  \end{tabular}
  \caption{
    Comparing our iteratively pretrained
    multilingual order-preserving lexiconized transformer (IPML)
    with the baselines training on 1,093 lines of English data
    in \textit{FAMO$^+$}. We checkmark the key components used
    in each experiments and explain all the baselines in
    details in Section \ref{results}.
  }
    \label{table:englishCompare}
\end{table}

\section{Results}\label{results}
We compare our iteratively pretrained multilingual
order-preserving lexiconized transformer (IPML)
with five baselines in Table~\ref{table:englishCompare}. 
\textit{MLc} is a baseline model of
multilingual order-preserving
lexiconized transformer
training on complete configuration;
in other words, we skip the first stage of
pretraining and train on the second stage
in Chapter \ref{step2} only.
\textit{MLs} is a baseline model
of multilingual order-preserving lexiconized
transformer training on star configuration;
in other words, we skip both steps of
pretraining and train on the final stage in
Chapter \ref{step3} only.
\textit{PMLc} is a baseline
model of pretrained multilingual
order-preserving lexiconized transformer
training on complete configuration;
in other words, we skip the final stage
of training after completing both stages of
pretraining.
\textit{PMLs} is a baseline model of
pretrained multilingual order-preserving lexiconized
transformer training on star configuration;
in other words, after the first stage of pretraining,
we skip the second stage
of pretraining and proceed to the final training
directly. Finally,
\textit{AML} is a baseline model of
multilingual order-preserving lexiconized transformer
on asymmetric data. We replicate the $\sim$1,000
lines of the low resource data
till it matches the training size of
other source languages; we train on
the complete graph configuration using 
eleven languages. Though the number of low resource
training lines is the same,
information is highly asymmetric. 

\begin{table}[t]
  \centering
  \small
  \begin{tabular}{p{1.1cm}p{0.65cm}|p{1.15cm}p{0.6cm}|p{1.15cm}p{0.6cm}}
    \toprule
    \multicolumn{6}{c}{Input Language Family} \\
    \midrule
    \multicolumn{2}{c|}{By Linguistics} & \multicolumn{2}{c|}{By Distortion} & \multicolumn{2}{c}{By Performance} \\
    \midrule
    \multicolumn{2}{c|}{\textit{FAMO$^+$}} & \multicolumn{2}{c|}{\textit{FAMD}} & \multicolumn{2}{c}{\textit{FAMP}} \\
    \midrule
    Source & BLEU & Source & BLEU & Source & BLEU \\
    \midrule
    Combined & 37.3 & Combined & 38.3 & Combined & 39.4 \\
    \midrule
    German & 35.0 & German & 36.7 & German & 37.6 \\
    Danish & 36.0 & Danish & 37.1 & Danish & 37.5 \\
    Dutch & 35.6 & Dutch & 35.6 & Dutch & 36.7 \\
    Norwegian & 35.7 & Norwegian & 36.9 & Norwegian & 37.1 \\
    Swedish & 34.5 & Afrikaans & 38.3 & Afrikaans & 39.3 \\
    Spanish & 36.4 & Marshallese & 34.7 & Spanish & 38.4 \\
    French & 35.3 & French & 36.0 & French & 36.6 \\
    Italian & 35.9 & Italian & 36.9 & Italian & 37.7 \\
    Portuguese & 31.5 & Portuguese & 32.9 & Portuguese & 33.1 \\
    Romanian & 34.6 & Frisian & 36.1 & Frisian & 36.9 \\
    \bottomrule
    \end{tabular}
  \caption{Performance of Iteratively Pretrained Multilingual Order-preserving Lexiconized Transformer (IPML)
    training for English on \textit{FAMO$^+$}, \textit{FAMD} and \textit{FAMP}. We train on only 1,093 lines of English data.}
    \label{table:english}
\end{table}

Pretraining is key as IPML beats the two baselines that
skip pretraining in Table~\ref{table:englishCompare}.
Using English as a hypothetical low resource language training on
FAMO$^+$, combined translation
improves from 13.4 (MLc) and 14.7 (MLs) to 37.3 (IPML)
with iterative pretraining. Training with the low resource
language on both the source and the target sides
boosts translation into the target side.
Star configuration has a slight
advantage over complete configuration as it
gives priority to translation
into the low resource language.
Iterative pretraining with BLEU score 37.3
has an edge over one stage of
pretraining with scores 34.7 (PMLc) and 35.7 (PMLs).  

All three pretrained models on symmetric data,
IPML, PMLc and PMLs, beat asymmetric baseline AML. 
In Table~\ref{table:englishCompare}, IPML has
a +10.3 BLEU increase over our asymmetric baseline
on combined translation using English
as a hypothetical low resource language training on FAMO$^+$.
All four use the same amount of data, but differ in
training strategies and data configuration.
In severely low resource scenarios, effective
training strategies on symmetric data improve translation greatly.

\begin{table}[t] 
  \centering
  \small
  \begin{tabular}{p{1.1cm}p{0.65cm}|p{1.15cm}p{0.6cm}|p{1.15cm}p{0.6cm}}
    \toprule
    \multicolumn{6}{c}{Input Language Family} \\
    \midrule
    \multicolumn{2}{c|}{By Linguistics} & \multicolumn{2}{c|}{By Distortion} & \multicolumn{2}{c}{By Performance} \\
    \midrule
    \multicolumn{2}{c|}{\textit{FAMO$^+$}} & \multicolumn{2}{c|}{\textit{FAMD}} & \multicolumn{2}{c}{\textit{FAMP}} \\
    \midrule
    Source & BLEU & Source & BLEU & Source & BLEU \\
    \midrule
    Combined & 23.0 & Combined & 23.1 & Combined & 22.2 \\
    \midrule
    Chuj & 21.8 & Chuj & 21.9 & Chuj & 21.6 \\ 
    Cakchiquel & 22.2 & Cakchiquel & 22.1 & Cakchiquel & 21.3 \\ 
    Guajajara & 19.7 & Guajajara & 19.1 & Guajajara & 18.8 \\
    Mam & 22.2 & Russian & 22.2 & Mam & 21.7 \\ 
    Kanjobal & 21.9 & Toba & 21.9 & Kanjobal & 21.4 \\ 
    Cuzco & 22.3 & Myanmar & 19.1 & Thai & 21.8 \\
    Ayacucho & 21.6 & Slovenský & 22.1 & Dadibi & 19.8 \\
    Bolivian & 22.2 & Latin & 21.9 & Gumatj & 19.1 \\
    Huallaga & 22.2 & Ilokano & 22.5 & Navajo & 21.3 \\
    Aymara & 21.5 & Norwegian & 22.6 & Kim & 21.5 \\
    \bottomrule
  \end{tabular}
  \caption{Performance of Iteratively Pretrained Multilingual Order-preserving Lexiconized Transformer (IPML)
    training for Eastern Pokomchi on \textit{FAMO$^+$}, \textit{FAMD} and \textit{FAMP}. We train on only 1,086 lines of Eastern Pokomchi data.}
    \label{table:pokomchi}
\end{table}

We compare IPML results training on different
sets of source languages in FAMO$^+$, FAMD, and FAMP,
for English and Eastern Pokomchi in
Table~\ref{table:english} and \ref{table:pokomchi}.
FAMP performs the best for translation into English while both 
FAMP and FAMD outperforms
FAMO$^+$ as shown in Table~\ref{table:english}.
FAMD performs best for translation into Eastern
Pokomchi as shown in Table~\ref{table:pokomchi}. Afrikaans has the
highest score for English's FAMD and FAMP, outperforming Dutch,
German or French. A reason may be that Afrikaans
is the youngest language in the Germanic family
with many lexical and syntactic borrowings from English
and multiple close neighbors of English \cite{gordon2005ethnologue}. 
When language family information is limited,
constructing FAMC to determine neighbors
is very useful in translation. 

\begin{table}[t]

\parbox{.45\linewidth}{
  \centering
  \small
  \begin{tabularx}{.45\columnwidth}{XX}
    \toprule
    Source & BLEU \\
    \midrule
    Combined & N.A. \\
    \midrule
    German & 34.8 \\
    Danish & 37.7 \\
    Dutch & 39.7 \\
    Swedish & 37.7 \\    
    Spanish & 42.8 \\ 
    French & 41.6 \\
    Italian & 39.2 \\
    Portuguese & 42.8 \\
    Romanian & 40.0 \\    
    Polish & 34.1 \\
    \bottomrule
    \end{tabularx}
  \caption{IPML Performance on the EMEA dataset trained on only 1,093 lines of English data.}
    \label{table:EMEA}
}
\hfill
\parbox{.45\linewidth}{
  \centering
  \small
  \begin{tabularx}{.45\columnwidth}{XX}
    \toprule
    Source & BLEU \\
    \midrule 
    Combined & 23.7 \\
    \midrule
    German & 21.6 \\
    Danish & 22.9 \\
    Dutch & 21.2 \\
    Norwegian & 21.3 \\
    Swedish & 19.9 \\
    Spanish & 22.9 \\
    French & 22.3 \\
    Italian & 21.8 \\
    Portuguese & 20.7 \\
    Romanian & 16.3 \\
    \bottomrule
    \end{tabularx}
  \caption{IPML Performance on the entire Bible excluding $\sim$1k lines of training and validation data.}
  \label{table:bible}
}
\end{table}

\begin{table*}[t]
    \small
    \centering
    \begin{tabularx}{\textwidth}{XXX}
      \toprule
      Source Sentence & IPML Translation & Reference \\ \midrule
      Caso detecte efeitos graves ou outros efeitos não mencionados neste folheto, informe o médico veterinário.
      &
      If you notice any side effects or other side effects not mentioned in this leaflet, please inform the vétérinaire.
      &
      If you notice any serious effects or other effects not mentioned in this leaflet, please inform your veterinarian. \\
      \midrule
      No tratamento de Bovinos com mais de 250 Kg de peso vivo, dividir a dose de forma a não administrar mais de 10 ml por local de injecção.
      &
      In the treatment of infants with more than 250 kg in vivo body weight, a the dose to not exceed 10 ml per injection.
      &
      For treatment of cattle over 250 kg body weight, divide the dose so that no more than 10 ml are injected at one site. \\
      \midrule
      No entanto, uma vez que é possível a ocorrência de efeitos secundários, qualquer tratamento que exceda as 1-2 semanas deve ser administrado sob supervisão veterinária regular.
      &
      However, because any of side effects is possible, any treatment that 1-5 weeks should be administered under regular supraveghere.
      &
      However, since side effects might occur, any treatment exceeding 1–2 weeks should be under regular veterinary supervision.
      \\
      \bottomrule
    \end{tabularx}
    \caption{Examples of IPML translation on medical EMEA dataset 
      from Portuguese to English using \textit{FAMO$^+$}. }
    \label{table:EMEAQualitative}
\end{table*}

Comparing Eastern Pokomchi results
with English results, we see that translation into real-life
severely low resource languages is more
difficult than translation into hypothetical ones.
The combined score is 38.3
for English in Table~\ref{table:english}
and 23.1 for Eastern Pokomchi on FAMD
in Table~\ref{table:pokomchi}. Eastern Pokomchi
has ejective consonants which makes
tokenization process difficult.
It is agglutinative, morphologically
rich and ergative just like Basque \cite{aissen2017mayan, clemens2015ergativity}. It is complex, unique and nontransparent to the
outsider \cite{england2011grammar}.
Indeed, translation into real severely
low resource languages is difficult. 

We are curious of how our model trained
on $\sim$1,000 lines of data performs on the rest of the Bible.
In other words, we would like to know how IPML performs
if we train on $\sim$3.5\% of the
Bible and test on $\sim$96.5\% of the Bible. In
Table~\ref{table:bible}, training on 1,093 lines from
the book of Luke,
we achieve a BLEU score of 23.7 for IPML
using FAMP in English \footnote{A previous version of this paper
shows higher BLEU scores with random sampling. Since
active learning is not the focus of this paper, 
we show all results training on the book of Luke in this paper.
For further results in active learning, please refer to our 
follow-up work \cite{zhou2021active}.}. 

We show qualitative examples 
in Table~\ref{table:englishQualitative}
and \ref{table:pokomchiQualitative}.
The source content is translated well
overall and there are a few places for improvement
in Table~\ref{table:englishQualitative}.
The words ``fishermen'' and ``fishers'' are paraphrases of
the same concept. IPML predicts
the correct concept though it is penalized by BLEU.

Infusing the order-preserving lexiconized component
to our training greatly improves qualitative evaluation. But it 
does not affect BLEU much as BLEU has its limitations
in severely low resource scenarios. This is why all
experiments include the lexiconized component in training. 
The BLEU comparison in our paper also
applies to the comparison of all experiments without the
order-preserving lexiconized component. This is important
in real-life situations when a low resource lexicon list
is not available, or has to be invented. For example, a
person growing up in a local village in Papua New Guinea
may have met many people named ``Bosai'' or ``Kaura'', but 
may have never met a person named ``Matthew'', 
and we may need to create a lexicon
word in the low resource language for ``Matthew''
possibly through phonetics.

We also see good results with the medical
EMEA dataset. Treating English as a hypothetical
low resource language, we train on only 1,093 lines
of English data. For Portuguese-English translation,
we obtain a BLEU score of 42.8 while the rest of
languages all obtain BLEU scores above 34 in
Table~\ref{table:EMEA} and Table~\ref{table:EMEAQualitative}.
In Table~\ref{table:EMEAQualitative}, we see that our
translation is very good, though a few words are
carried from the source language including ``vétérinaire''.
This is mainly because our $\sim$1,000 lines contain
very small vocabulary; however, by carrying the
source word over, key information is preserved. 

\section{Conclusion}\label{conclusion}
We use $\sim$1,000 lines of low resource data
to translate a closed text that is known in advance to
a severely low resource language by leveraging massive
source parallelism.
We present two metrics to rank the
124 source languages and construct FAMCs.  
We build an
iteratively pretrained multilingual
order-preserving lexiconized transformer and
combine translations from all source languages
into one by using our centric measure.
Moreover, we add a multilingual order-preserving lexiconized 
component to translate the named entities
accurately. We build a massively parallel lexicon table for
2,939 Bible named entities in 124 source languages,
covering more than 66 severely low resource languages.
Our good result for the medical EMEA dataset shows that
our method is useful for other datasets and applications. 

Our final result can also serve as a ranking
measure for linguistic distances
though it is much more expensive
in terms of time and resources.
In the future, we would like to explore more metrics that are
fast and efficient in ranking linguistic distances to
the severely low resource language.

%


\bibliographystyle{acl_natbib}
\interlinepenalty=10000
\bibliography{thesis}

\newpage\hbox{}\thispagestyle{empty}\newpage
\pagebreak
\section{Appendix}
In Table~\ref{table:ranking} and Table~\ref{table:pokomchi}, Kanjobal is
Eastern Kanjobal, Mam is Northern Mam, Cuzco is Cuzco Quechua, Ayacucho is Ayacucho Quechua,
Bolivian is South Bolivian Quechua, and Huallaga is Huallaga Quechua.
  
\begin{figure*}[ht!]
  \centering
  \includegraphics[width=0.6\linewidth]{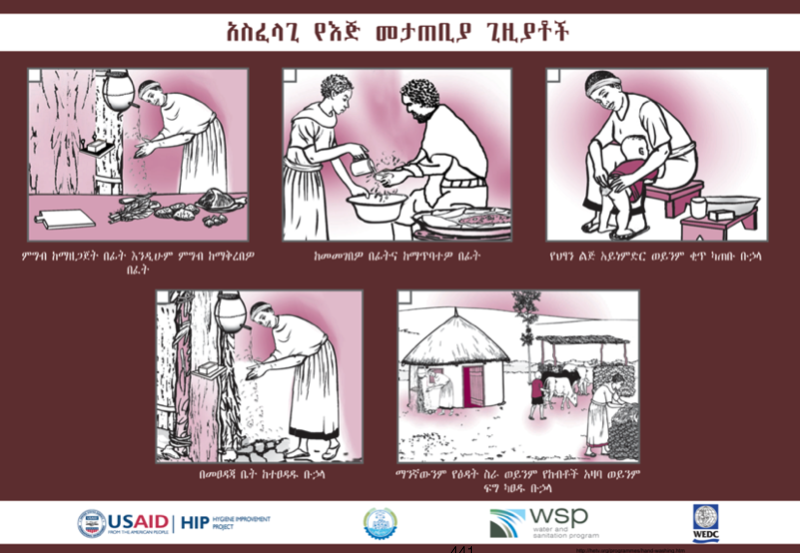}
  \caption{An Amharic illustration of translation of water, sanitation,
and hygiene (WASH) guidelines in Ethiopia \cite{wash}. }
  \label{fig:wash}
\end{figure*}

We show an illustration of WASH guidelines in Figure~\ref{fig:wash}.
We also show IPML translations into Eastern Pokomchi (Mayan) in Table~\ref{table:pokomchiQualitative}. 
\begin{table*}[t]
    \small
    \centering
    \begin{tabular}{p{4.5cm}p{4.6cm}p{5cm}}
      \toprule
      Source Sentence & IPML Translation & Reference \\ \midrule
      Ket idi limmabas iti dinna ti baybay ti Galilea, nakitana ni Simon ken ni Andres a cabsatna, nga iwaywayatda ti iket iti baybay; ta dumadaclisda idi.
      &
      Eh noq ojik i rub{\textquotesingle}an i Jesús juntar i k{\textquotesingle}isa palaw i Galilea, xrilow reje i Simón ruch{\textquotesingle}ihil i Andres, re{\textquotesingle} i rutuut i k{\textquotesingle}isa palaw, ruum jinaj i k{\textquotesingle}isa palaw barco.
      &
      Noq k{\textquotesingle}ahchi{\textquotesingle} rik{\textquotesingle}iik i Jesús chi chii{\textquotesingle} i k{\textquotesingle}isa palaw ar Galilea, xrilow reje wach i Simón ruch{\textquotesingle}ihil i ruchaaq{\textquotesingle}, Andres rub{\textquotesingle}ihnaal. Re{\textquotesingle} keh aj karineel taqe, k{\textquotesingle}ahchi{\textquotesingle} kikutum qohoq i kiya{\textquotesingle}l pan palaw.
      \\ \midrule
      Ket idi nagna pay bassit nakitana ni Santiago nga anac ni Zebedeo ken ni Juan a cabsatna, nga addada idi iti barangayda, a tartarimaanenda dagiti iketda.
      &
      Eh noq ojik i rub{\textquotesingle}an i Jesús, xrilow i Jacobo, re{\textquotesingle} i Jacobo rak{\textquotesingle}uun i Zebedeo, re{\textquotesingle} Juan rub{\textquotesingle}ihnaal, ruch{\textquotesingle}ihil taqe i raj tahqaneel. eh xkikoj wo{\textquotesingle} wach chinaah i k{\textquotesingle}isa palaw.
      &
      Eh junk{\textquotesingle}aam-oq chik i xb{\textquotesingle}ehik reje i Jesús, xrilow kiwach i ki{\textquotesingle}ib{\textquotesingle} chi winaq kichaaq{\textquotesingle} kiib{\textquotesingle}, re{\textquotesingle} Jacobo, re{\textquotesingle} Juan, rak{\textquotesingle}uun taqe i Zebedeo. Eh wilkeeb{\textquotesingle} chupaam jinaj i barco, k{\textquotesingle}ahchi{\textquotesingle} kik{\textquotesingle}ojem wach i kiya{\textquotesingle}l b{\textquotesingle}amb{\textquotesingle}al kar.
      \\ \midrule
      Ket immasideg ni Jesus ket iniggamanna iti imana ket pinatacderna; ket pinanawan ti gorigor , ket nagservi cadacuada.
      &
      Eh re{\textquotesingle} Jesús xujil i koq riib{\textquotesingle}, xutz{\textquotesingle}a{\textquotesingle}j i koq chinaah i q{\textquotesingle}ab{\textquotesingle}. eh re{\textquotesingle} i kaq tz{\textquotesingle}a{\textquotesingle} chi riij. eh jumehq{\textquotesingle}iil xwuktik johtoq, re{\textquotesingle} chik i reh xutoq{\textquotesingle}aa{\textquotesingle} cho yej-anik kiwa{\textquotesingle}.
      &
      Eh re{\textquotesingle} i Jesús xujil i koq riib{\textquotesingle} ruuk{\textquotesingle} i yowaab{\textquotesingle}, xuchop chi q{\textquotesingle}ab{\textquotesingle}, xruksaj johtoq, eh jumehq{\textquotesingle}iil xik{\textquotesingle}ik i tz{\textquotesingle}a{\textquotesingle} chi riij. Eh re{\textquotesingle} chik i reh xutoq{\textquotesingle}aa{\textquotesingle} cho yej-anik kiwa{\textquotesingle}.
      \\ \bottomrule
    \end{tabular}
    \caption{Examples of Iteratively Pretrained Multilingual
      Order-preserving Lexiconized Transformer (IPML) translation
      from Ilokano to Eastern Pokomchi using \textit{FAMD}. We
      train on only 1,086 lines of Eastern Pokomchi data.}
    \label{table:pokomchiQualitative}
  \end{table*}

\end{document}